\documentclass[letterpaper, 10 pt, conference]{ieeeconf}
\IEEEoverridecommandlockouts                              

\makeatletter
\let\NAT@parse\undefined
\makeatother

\usepackage{graphics} 
\usepackage{subcaption} 
\usepackage{epsfig} 

\usepackage{times}
\usepackage{mathtools} 
\usepackage{amssymb}  
\usepackage{color}
\usepackage[numbers]{natbib}
\usepackage{multicol}
\usepackage[bookmarks=true]{hyperref}
\usepackage{booktabs}
\usepackage{floatrow}
\usepackage{verbatim}
\usepackage{xcolor}

\newcommand\gdkcomment[1]{\textcolor{blue}{#1}}

\mathtoolsset{showonlyrefs,showmanualtags}

\DeclareMathOperator*{\argmin}{arg\,min}
\DeclareMathOperator*{\argmax}{arg\,max}

\begin{document}
\title{Hybrid Bayesian Eigenobjects: Combining Linear Subspace and Deep Network Methods for 3D Robot Vision}

\author{Benjamin Burchfiel$^\dagger$ and George Konidaris$^\star$\\
$\dagger$Duke University, Durham NC\\
$\star$Brown University, Providence RI\\
bcburch@cs.duke.edu, gdk@cs.brown.edu}

\maketitle

\begin{abstract}
We introduce Hybrid Bayesian Eigenobjects (HBEOs), a novel representation for 3D objects designed to allow a robot to jointly estimate the pose, class, and full 3D geometry of a novel object observed from a single viewpoint in a single practical framework. By combining both linear subspace methods and deep convolutional prediction, HBEOs  efficiently learn nonlinear object representations without directly regressing into high-dimensional space. HBEOs also remove the onerous and generally impractical necessity of input data voxelization prior to inference. We experimentally evaluate the suitability of HBEOs to the challenging task of joint pose, class, and shape inference on novel objects and show that, compared to preceding work, HBEOs offer dramatically improved performance in all three tasks along with several orders of magnitude faster runtime performance.
\end{abstract}

\IEEEpeerreviewmaketitle

\section{Introduction}
\label{sec:intro}

If we desire robots capable of functioning in arbitrary home and office settings---environments that exhibit huge amounts of variation---we require dramatic improvement to our perceptual systems. It is no longer feasible to rely on operating in known settings with previously encountered objects; generally-intelligent robots 
require
the ability to generalize from objects they have previously observed to similar but novel objects. Prior work has identified three critical inference capabilities required by most robots interacting with objects: classification, pose estimation, and 3D completion \cite{Burchfiel-RSS-17}. For a single observed object, that entails determining the object's type, its  position in the world, and its full 3D geometry. 
These capabilities are critical for many common robot tasks and constitute the perceptual foundation upon which complex object-centric behavior may be built. For instance, a household robot should be able to reason about a new and somewhat oddly shaped coffee mug, recognizing it as a coffee cup, inferring that it likely has a handle protruding from the rear (even if that portion is unseen), and estimating its pose.

This work presents a novel framework for object representation: Hybrid Bayesian Eigenobjects (HBEOs). HBEOs, like BEOs \cite{Burchfiel-RSS-17} which preceded them, are designed to generalize knowledge from previously encountered objects to novel ones. HBEOs are trained from a set of 3D meshes with known class and pose; they allow a novel object---observed via a depth image from a single viewpoint---to have its class, pose, and full 3D shape estimated.

HBEOs employ Variational Bayesian Principal Component Analysis (VBPCA) to learn a linear subspace in which objects lie. The critical insight is that the space spanned by all possible 3D structures is far larger than the---still quite sizable---space spanned by 3D objects encountered in the everyday world. By explicitly learning a compact basis for objects we wish our robot to reason about, we constrain the inference problem significantly. This representation allows a 3D object to be expressed as a low dimensional set of VBPCA coefficients, facilitating efficient pose estimation, classification, and completion without operating directly in high-dimensional object space. HBEOs use a learned non-linear method---specifically, a deep convolutional network \cite{lecun1995convolutional}---to determine the correct projection coefficients for a novel partially observed object. By combining linear subspace methods with deep convolutional inference, HBEOs draw from the strengths of both approaches.

Previous work on 3D shape completion employed either deep architectures which predict object shape in full 3D space (typically via voxel output) \cite{wu20153d, wu2016learning, dai2017complete, varlay2017} or linear methods which learn linear subspaces in which objects tend to lie \cite{Burchfiel-RSS-17}.  Both approaches have had some success, but
also have significant weaknesses. End-to-end deep methods suffer from the high dimensionality of object space; the data and computation requirements of regressing into $50,000$ or even million dimensional space are severe. Linear approaches, on the other hand, perform their predictions in an explicit low dimensional subspace. This approach is fast and quite data efficient but requires partially observed objects be voxelized before inference can occur, and lacks the expressiveness of a non-linear deep network.
Unlike existing approaches, which are either fully linear or do not leverage explicit object spaces, HBEOs have the flexibility of nonlinear methods without requiring expensive regression directly into high-dimensional space. Additionally, because HBEOs perform inference directly from a depth image, they do not require voxelizing a partially observed object, a process which requires estimating a partially observed object's full 3D extents and pose prior to voxelization. Empirically, we show our hybrid approach outperforms competing methods when performing joint pose estimation, classification, and 3D completion of novel objects.

\section{Background}
\label{sec:background}
\subsection{3D Object Representations}
The most common 3D object representations are meshes, voxel objects, multiview depth images, and pointclouds. Voxels (volumetric pixels) are expressive and benefit from being a fixed-size representation, but have cubic complexity with respect to their linear resolution and are inefficient for objects with regular geometry. Object meshes, which model just the surface of an object, efficiently capture fine detail but can be cumbersome to work with due to their variable size and lack of explicit volume representation. It can also be difficult to generate complete meshes from imperfect or noisy data. Multiview representations offer efficiency advantages compared to voxel approaches and ease of generation compared to mesh models, but do not explicitly represent object interiors. Pointclouds are often convenient to generate and can be very expressive, but do not explicitly model free and occupied space and are not a fixed size. Historically, depth images and multiview representations have been popular for 3D classification while meshes, voxels, and pointclouds are more common when performing object manipulation \cite{Rusu20113DIH, carbone2012grasping}.

While most modern object-centric perceptual systems use convolutional networks at their core, other methods of performing inference on 3D objects have been created over the years for various tasks. Parts-based methods create a parts dictionary and represent objects as a composition of multiple dictionary entries \cite{3D_Parts, marini2006partial}, while database methods, common in robotics, construct an explicit database consisting of known objects and perform various types of inference via queries into that database \cite{guibas_database_objects, guibas_database_objects2, guided3DScanning}.

\subsection{3D Classification}
Object classification, the task of determining an object's semantic ``type'', is a common perceptual task in robotics. 2D classification is  extremely well studied  \cite{krizhevsky2012imagenet, NIPS2010_4064,gehler2009feature} and in recent years 3D classification has gained considerable research attention. Modern approaches to 3D object classification generally rely on convolutional deep networks which are trained from full 3D models of objects. Broadly speaking, most current 3D classification approaches fall into two categories: multiview and volumetric. Volumetric approaches explicitly represent 3D objects as volumes (generally via voxels) \cite{maturana2015voxnet,Qi_2016_CVPR, 7273863,wu20153d} while multiview approaches represent a 3D object as a series of 2D or 2.5D views \cite{su2015multi, bai2016gift}. 
Both methods have shown promising results and research has been conducted to combine the two representations \cite{hegde2016fusionnet}. 

\subsection{3D Pose Estimation}
Pose estimation is the task of determining an object's position and orientation in the world. With rigid objects, pose estimation generally consists of either a three degree of freedom orientation estimate or a six degree of freedom orientation and position estimate. Most methods apply to single-instance pose estimation: a known object with unknown pose. In these settings, when a full 3D model of the object is available, techniques such as Iterative Closest Points \cite{ICP}, Bingham Distribution Pose Estimation \cite{Glover-RSS-11}, and DNN approaches \cite{tulsiani2015viewpoints} have shown to be effective. Nevertheless, none of these methods are designed to predict the pose of a novel object relative to a canonical class-specific baseline pose. One method that has shown promise is proposed by \citet{elhoseiny2015convolutional}; they developed a convolutional network capable of predicting one degree of freedom pose (into sixteen discrete bins) for a novel object. More recently, \citet{Burchfiel-RSS-17} propose a representation capable of performing up to three degree of freedom pose estimation on novel objects, but computational performance remains a bottleneck due to their pose estimation by search approach.

\subsection{3D Completion}
3D object completion consists of inferring the full 3D geometry of a partially observed 3D object. Initial work in this area focused on model repair; objects were mostly observed but contained holes and other small patches of missing information. By exploiting surrounding geometry and symmetry properties, these approaches could repair small gaps but were not designed to patch large missing portions of objects  \cite{attene2010lightweight}. More recently, new symmetry exploitation methods have been proposed which rely on global-scale object symmetry to estimate the shape of unobserved object regions \cite{schiebener2016heuristic, Song_2016_CVPR}. Unlike the initial hole filling work, these symmetry methods can reconstruct large missing portions of objects but fail when when their symmetry heuristic is not applicable. Database methods have also been popular, particularly in robotics. These methods treat a partially observed object as a query into a database of known (complete) 3D objects  \cite{guibas_database_objects, guibas_database_objects2}. When a close match is found they 
perform 
well. Unfortunately---due to the huge amount of object variation in household environments---database methods have scaling issues and perform poorly when a close match is not found. Hybrid approaches have also been proposed that attempt to perform shape completion locally and use the completed object as a database query to obtain a final estimate \cite{dai2017complete}, but these methods require 
a large and dense object model database.

The current state of the art in 3D object completion consists of approaches in two categories: deep regression methods and linear subspace methods. They generally take as input a partially observed voxel object and output a completed (fully specified) version. Deep approaches train convolutional networks to estimate either surface normals \cite{DBLP:journals/corr/TulsianiKHCM15} or full 3D structure of partially observed objects \cite{wu20153d, dai2017complete, Soltani2017Synthesizing3S}. While these methods are quite powerful, they tend to be data inefficient and can struggle with producing high-resolution output, although \citet{dai2017complete} attempt to address this issue by maintaining an explicit database of known high-resolution objects. Linear subspace methods have the advantage of operating in an explicitly constructed low-dimensional space and have empirically demonstrated competitive performance with deep approaches \cite{Burchfiel-RSS-17}. These methods learn a subspace on which common objects lie and perform inference in that subspace. \citet{Burchfiel-RSS-17} treat shape completion as an under-constrained forward projection problem; given some partially observed object, $\mathbf{\hat{o}}$, find a \textit{good} projection onto the learned low-dimensional subspace.

\subsection{Variational Bayesian Principal Component Analysis}
Similarly to \citet{Burchfiel-RSS-17}, our work employs Variational Bayesian Principal Components Analysis (VBPCA) \cite{bishop1999variational} to learn an explicit low-dimensional object space. VBPCA is a fully Bayesian version of probabilistic PCA (PPCA) \cite{PPCA}, itself a probabilistic extension of traditional linear Principal Component Analysis (PCA). PPCA represents each datapoint,
$\mathbf{x_i}$, as
\begin{equation}
\mathbf{x_i} = \mathbf{W}\mathbf{c_i} + \boldsymbol{\mu} + \epsilon_i,
\end{equation}
where $\mathbf{X}$ is a data matrix where column $i$ of $\mathbf{X}$ is $\mathbf{x_i}$, $\mathbf{W}$ is a  basis matrix, $\mathbf{c_i}$ are the projection coefficients of $\mathbf{x_i}$ onto  $\mathbf{W}$, $\boldsymbol{\mu}$ is the mean datapoint, and $\mathbf{\epsilon_i}$ is zero-mean Gaussian noise for point $i$. PPCA's parameters are typically estimated via Expectation-Maximization (EM) by alternating between updating the projection coefficients for each datapoint, $\mathbf{c_i}$, and updating $\mathbf{W}$, $\boldsymbol{\mu}$, and $\epsilon$.

Bayesian PCA (BPCA) \cite{bpca1} places (Gaussian) priors, $\mathcal{H}$, over $W$ and $\boldsymbol{\mu}$, allowing BPCA to model the entire posterior probability of model parameters:
\begin{equation}
\label{eq:1}
p(\mathbf{W}, \boldsymbol{\mu}, \mathbf{C}| \mathbf{X}, \mathcal{H}).
\end{equation}
VBPCA overcomes the intractability of evaluating \eqref{eq:1} by (approximately) factorizing the posterior:
\begin{equation}
q(\mathbf{W}, \boldsymbol{\mu}, \mathbf{C}) = \prod_{i=1}^dq(\mathbf{\mu_i})\prod_{i=1}^dq(\mathbf{w}_i)\prod_{i=1}^nq(\mathbf{c_i}),
\end{equation}
allowing factors to be updated individually during EM \cite{bishop1999variational}. Bishop \cite{bishop1999variational} demonstrates that the regularizing nature of VBPCA's priors make it far more suitable than PCA when the number of datapoints ($n$) being learned from is significantly smaller than the dimensonality of the original space ($d$). For object representation, where voxel space is typically on the order of tens or hundreds of thousands of dimensions and $n \ll d$, VBPCA offers significant advantages over PCA \cite{Burchfiel-RSS-17}.

\subsection{Bayesian Eigenobjects}
Bayesian Eigenbojects (BEOs) 
\cite{Burchfiel-RSS-17} are a unified framework for object representation. The key to the usefulness of BEOs lies in a learned low-dimensional subspace in which common objects lie. BEOs use VBPCA to learn a class-specific subspace for each class and then construct a single shared subspace that is well suited to represent all training classes. Let $\mathbf{W_i}$ and $\boldsymbol{\mu_i}$ represent the VBPCA-learned subspace for class $i$. The shared subspace, $\mathbf{W}$, is
\begin{equation}
\mathbf{W} = [\mathbf{W_1},...,\mathbf{W_m}, \boldsymbol{\mu_1},...,\boldsymbol{\mu_m}],
    \label{eq:bigSubspace}
\end{equation}
where $\mathbf{W}$ is a $d\times2m$ matrix with rows corresponding to dimensions in voxel space and columns that form basis vectors for representing objects in the low-dimensional subspace. Without loss of generality (i.e. expressive power) we further simplify $\mathbf{W}$ by finding an orthonormal basis spanning the columns of $\mathbf{W}$. Let $\mathbf{W'} = \mathbf{U'}$ where $\mathbf{U}\mathbf{S}\mathbf{V^T} = \mathbf{W}$ is the singular value decomposition (SVD) of $\mathbf{W}$ and $\mathbf{U'}$ is the first $rank(\mathbf{W})$ rows of $\mathbf{U}$. For simplicity, we refer to this orthonormal basis, $\mathbf{W'}$, as $\mathbf{W}$ for the remainder of the paper.

BEOs treat (voxelized) objects as points in $d$-dimensional voxel space by unraveling voxel objects into $d$-dimensional vectors. $\mathbf{o}$, a novel object, can be projected onto $\mathbf{W}$ via
\begin{equation}
\mathbf{o}'=\mathbf{W}^T\mathbf{o}.
\label{eq:project_complete}
\end{equation}
Furthermore, any point which has been projected onto $\mathbf{W}$ may be \textit{back-projected} (converted back into voxel-object form) by
\begin{equation}
\mathbf{\hat{o}}= \mathbf{W}\mathbf{o}'.
\label{eq:reconst}
\end{equation}
$\mathbf{\hat{o}}$ is generally referred to as the ``completed'' or ``reconstructed'' version of $\mathbf{o}$, and $\mathbf{o}'$ as the ``projected'' version of $\mathbf{o}$.\footnote{After applying Equation \eqref{eq:project_complete}, values will generally no longer be exactly $0$ or $1$. In our work, we threshold at $0.5$ to rebinarize the completed object.}

The key to BEO's usefulness is partial-projection: the ability to project a partially observed object onto the subspace. Given a partially observed voxel object consisting of known-filled, known-empty, and unobserved voxels, BEOs seek to minimize the reconstruction error with respect to only the known-filled and known-empty portions of the object. Let $V$ be an object-specific $d'$ by $d$ binary selection matrix such that $\mathbf{V}\mathbf{o}=\mathbf{w}$, where $\mathbf{w}$ is a length $d'<d$ vector consisting of the known elements of $\mathbf{o}$. The error induced by an arbitrary projection of $\mathbf{o}$ is
\begin{equation}
E(\mathbf{o'}) = ||V(\mathbf{W}\mathbf{o'}) - \mathbf{w}||^2_2
\end{equation}
and the gradient of this error is
\begin{equation}
2\mathbf{W}^TV^T[V(\mathbf{W}\mathbf{o'}) - \mathbf{w}].
\end{equation}
We can estimate $\mathbf{o'}$ via
\begin{figure*}[ht!]
\centering
\includegraphics[width= \textwidth]{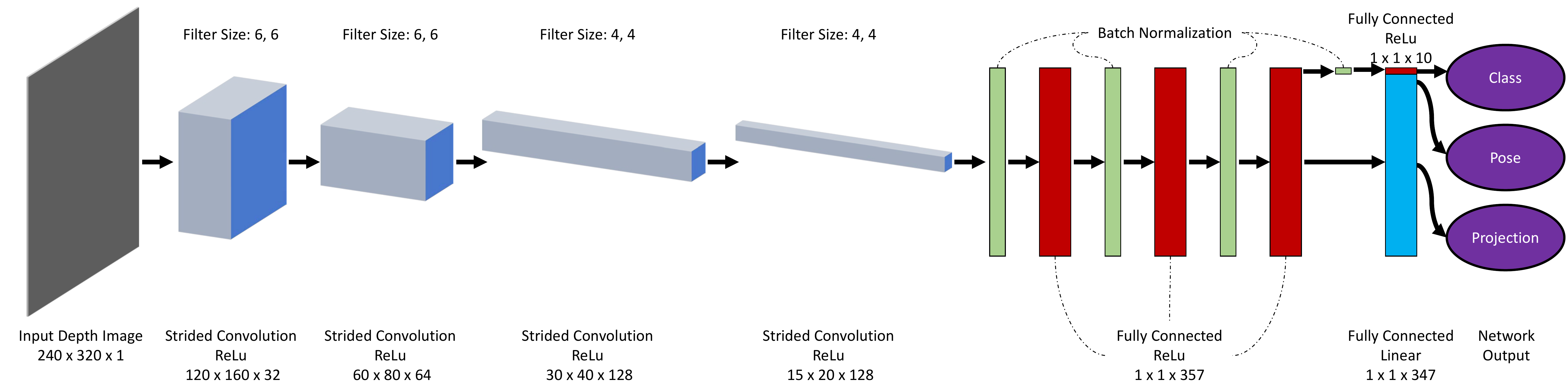}
\caption{The architecture of HBEONet (containing approximately 15 million parameters). Each convolution has a stride of 2x2 and pooling is not used. The non-trained softmax layer applying to the class output is not pictured.}
\label{fig:HBEONet}
\end{figure*}

\begin{equation}
\mathbf{A}\mathbf{o'} =\mathbf{b},
\label{eq:projIncomplete}
\end{equation}
or in the regularized case
\begin{equation}
\mathbf{o'} = \argmin_{\mathbf{o'}}~||\mathbf{A}\mathbf{o'} - \mathbf{b}||_2 + \lambda||\mathbf{o'}||_1\;,
\label{eq:lassoProj}
\end{equation}
where
\begin{equation}
\mathbf{A} = \mathbf{W}^T \mathbf{V}^T \mathbf{V} \mathbf{W}
\end{equation}
and
\begin{equation}
\mathbf{b} = \mathbf{W}^T \mathbf{V}^T\mathbf{w}.
\end{equation}
Once a projection into the learned subspace is found, back-projection (to obtain an estimate of the object's full 3D geometry) can proceed as normal via equation \eqref{eq:reconst}. To perform classification and pose estimation in this subspace, BEOs learn a Gaussian mixture model (GMM) over object class and pose from the training data. When a novel object is encountered, it is perturbed into multiple candidate orientations, each candidate is projected onto the subspace, and the most likely class-pose pair is estimated via
\begin{equation}
\{\hat{r}, \hat{c}\} = \argmax_{r\in R,~c \in C}~~\frac{P(r)D(\mathbf{o^{'r}} | c)P(c)}{\sum_{r_j \in R}\sum_{c_j\in C}P(r_j)D(\mathbf{o^{'{r_j}}} | c_j)P(c_j)},
\label{eq:densityRotation2}
\end{equation}
where $P(r)$ is the prior probability of rotation $r$, $D(\mathbf{o^{'r}} | c)P(c)$ is the probability of the completed object conditioned on class $c$, $P(c)$ is the prior probability of class $c$, $C$ is a set of classes, $\mathbf{o^r}$ denotes the object rotated by $r$, and $R$ is a set of rotations. While BEOs exhibit reasonable performance when estimating 1-DOF pose, this process becomes prohibitively expensive for additional degrees of freedom. Furthermore, BEOs assume that a partially observed object with unknown pose can be properly (and consistently) voxelized. In practice, this voxelization is challenging and often infeasable, relying on aligning a small portion of a novel object correctly in 7 degrees of freedom (orientation, translation, and scale).

\section{Hybrid Bayesian Eigenobjects}
\begin{figure*}[ht!]
\centering
\includegraphics[width= 1.0\textwidth]{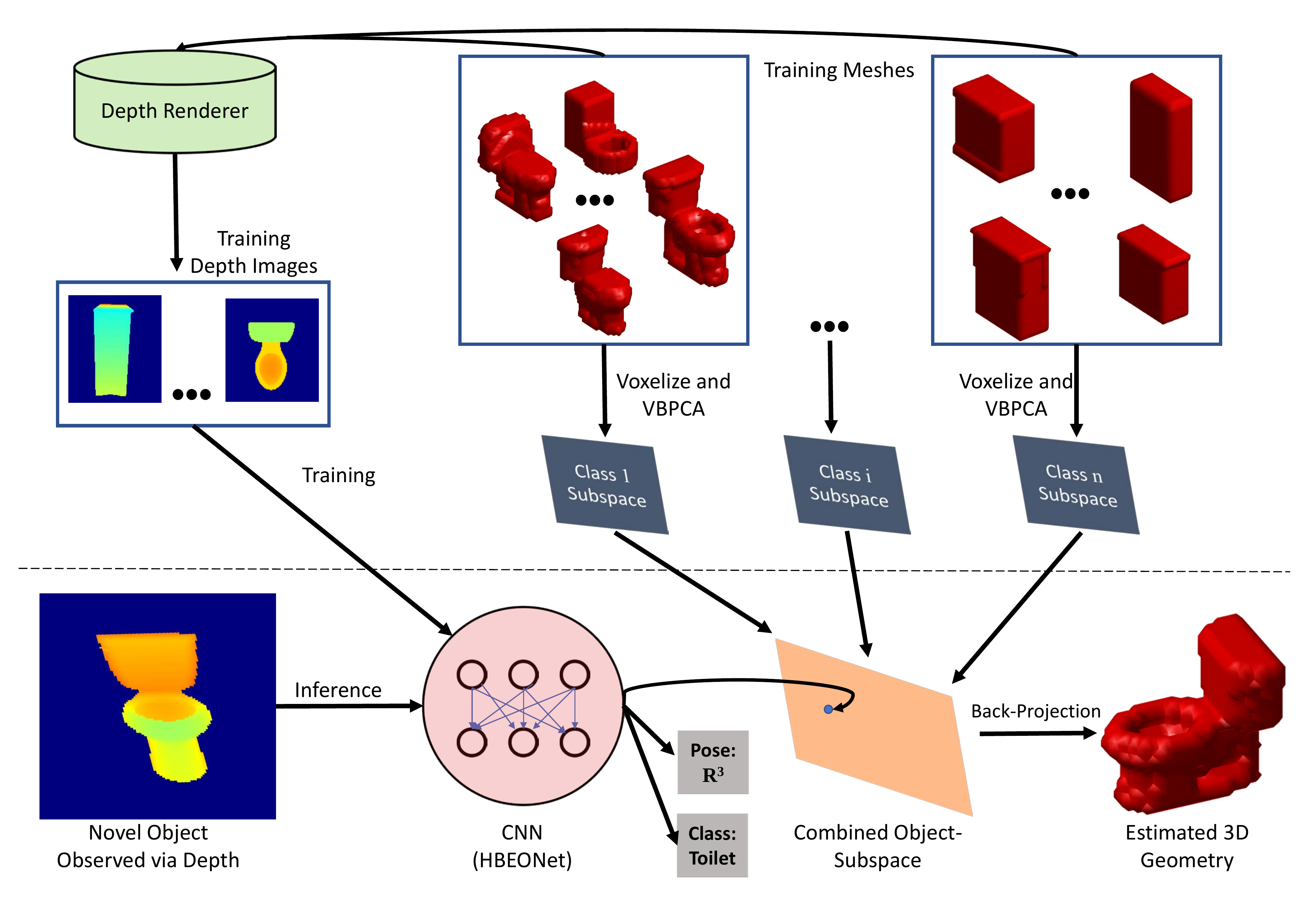}
\caption{Overview of the HBEO framework. Portions above the dotted line correspond to training operations while the bottom area denotes inference.}
\label{fig:pipeline}
\end{figure*}
HBEOs use an internal voxel representation, similar to both \citet{wu20153d} and \citet{Burchfiel-RSS-17}, but operate over raw depth images as input, avoiding the onerous requirement of voxelizing input at inference time. Like BEOs, HBEOs learn a single shared object-subspace; however, HBEOs learn a mapping directly from depth input into the learned low-dimensional subspace and predict class and pose simultaneously, allowing 1-shot pose, class, and shape estimation. Once HBEOs have projected an object into the learned subspace, the object's full 3D geometry may be estimated via equation \eqref{eq:reconst}. HBEOs:
\begin{enumerate}
  \item Operate directly on (segmented) depth images.
  \item Use a learned non-linear mapping (HBEONet) to project novel objects onto an object subspace.
  \item Predict the subspace projection jointly with class and pose in a single shot.
\end{enumerate}

\subsection{Learning a Projection into the Subspace}
HBEOs use a partially convolutional deep-network (HBEONet) to jointly predict class, pose, and a projection into the low-dimensional subspace given a depth image. HBEONet consists of four shared strided convolutional layers followed by three shared fully connected layers with a final separated layer for classification, pose estimation, and subspace projection. Figure \ref{fig:HBEONet} provides an overview of HBEONet's structure. This shared architecture incentivises the predicted class, pose, and predicted 3D geometry to be mutually consistent and ensures that learned low-level features are useful for multiple tasks. In addition to being fast, HBEOs leverage much more nuanced information during inference than BEOs. When BEOs perform object completion via \eqref{eq:lassoProj}, each piece of object geometry is equally important. A voxel representing the side of a toilet, for instance, is weighted equivalently to a voxel located in the toilet bowl. In reality however, certain portions of geometry contain far more information than others. Observing a portion of toilet bowl is far more informative than observing a piece of geometry on the side of the tank. HBEONet is able to learn these relationships: some learned features may be far more germane to the estimated output than others, providing a significant performance increase. Because HBEONet predicts subspace-projections instead of directly outputting 3D geometry (like end-to-end deep approaches), it need only produce several hundred dimensional output instead of regressing into tens or hundreds of thousands of dimensions. HBEOs
thereby combine appealing elements of both deep inference and efficient subspace techniques.

\subsection{Input-Output Encoding and Loss}
HBEOs take a single pre-segmented depth image (such as that produced via a Kinect or RealSense sensor) at inference time and produce three output predictions: A subspace projection (a vector in $\mathbb{R}^d$), a class estimate (via softmax), and a pose estimate (via three element axis-angle encoding).

The loss function used for HBEONet is
\begin{equation}
\mathcal{L} = \gamma_c\mathcal{L}_c + \gamma_o\mathcal{L}_o + \gamma_p\mathcal{L}_p,
\end{equation}
where $\mathcal{L}_c$, $\mathcal{L}_o$, and $\mathcal{L}_p$ represent the classification, orientation, and projection losses (respectively) and $\gamma_c$, $\gamma_o$, and $\gamma_p$ weight the relative importance of each loss. Both $\mathcal{L}_O$ and $\mathcal{L}_P$ are given by Euclidean distance between the network output and target vectors while $\mathcal{L}_C$ is obtained by applying a soft-max function to the network's classification output and computing the cross-entropy between the target and soft-max output. During training, depth-target pairs are provided to the network in the traditional supervised fashion. Figure \ref{fig:pipeline} illustrates the complete training and inference pipeline used in HBEOs.

\section{Experimental Evaluation}
\label{sec:experiments}
We evaluated the performance of HBEOs using the ModelNet10 dataset \cite{wu20153d} which consists of $4889$ common household object across 10 classes: \{Bathtubs, Beds, Chairs, Desks, Dressers, Monitors, Night Stands, Sofas, Tables, Toilets\}. ModelNet10 objects are provided in mesh form, are aligned to a standard pose, and are scaled to be a consistent size. While ModelNet10 has been widely used as a dataset, most of this use has been for benchmarking 3D classification from fully observed objects. Here, we employ ModelNet10 to evaluate performance on partially observed (single-viewpoint) objects in the challenging setting of joint pose, class, and 3D geometry estimation.

To obtain a shared object basis, each object mesh in ModelNet10 was voxelized to size $d=30^3$ and then converted to vector form (i.e. each voxel object was reshaped into a $27,000$ dimensional vector). VBPCA was performed separately for each class to obtain 10 class specific subspaces, each with basis size automatically selected to capture 60 percent of variance in the training samples (equating to between 30 and 70 retained components per class). We also employed zero-mean unit-variance Gaussian distributions as regularizing hyperparameters during VBPCA. After VBCPA, the class specific subspaces were combined via equation \eqref{eq:bigSubspace} into a single shared subspace with $344$ dimensions.

\begin{table*}[ht!]
\centering
  \begin{tabular}{lcccccccccc|c}  
  \toprule
    \textbf{Known Pose} & Bathtub & Bed & Chair & Desk & Dresser & Monitor & Night Stand & Sofa & Table & Toilet & \textbf{Total}\\
    \midrule
    BEO  \cite{Burchfiel-RSS-17} &$48.0$ &$\mathbf{95.0}$ &$\mathbf{93.0}$ &$\mathbf{46.5}$ &$64.0$ &$\mathbf{91.0}$ &$\mathbf{55.8}$ &$\mathbf{92.0}$ &$\mathbf{75.0}$ &$80.0$ &$\mathbf{76.3}$\\
    3DShapeNets \cite{wu20153d} &$\mathbf{76.0}$ &$77.0$ &$38.0$ &$22.1$ &$\mathbf{90.7}$ &$74.0$ &$38.4$ &$57.0$ &$1.0$ &$79.0$ &$54.4$\\
    \midrule
    \textbf{Unknown Pose} & Bathtub & Bed & Chair & Desk & Dresser & Monitor & Night Stand & Sofa & Table & Toilet & \textbf{Total}\\
    \midrule
    BEO \cite{Burchfiel-RSS-17} (1-DOF)&$4.0$ &$64.0$ &$83.0$ &$16.3$ &$51.2$ &$86.0$ &$36.0$ &$49.0$ &$\mathbf{76.0}$ &$46.0$ &$54.5$\\
    HBEO (3-DOF)&$\mathbf{91.3}$ &$\mathbf{86.4}$ &$\mathbf{84.1}$ &$\mathbf{57.6}$ &$\mathbf{79.7}$ &$\mathbf{97.9}$ &$\mathbf{81.3}$ &$\mathbf{75.4}$ &$72.3$ &$\mathbf{92.3}$ &$\mathbf{81.8}$\\
    \bottomrule
  \end{tabular}
  \caption{ModelNet10 classification accuracy (percent) with single-viewpoint queries comparing HBEOs with 3DShapeNets and BEOs.}
\label{table:partialClassification}
\end{table*}
\begin{figure*}[ht!]
\centering
\begin{subfigure}[t]{0.32\textwidth}
\includegraphics[width =\textwidth]{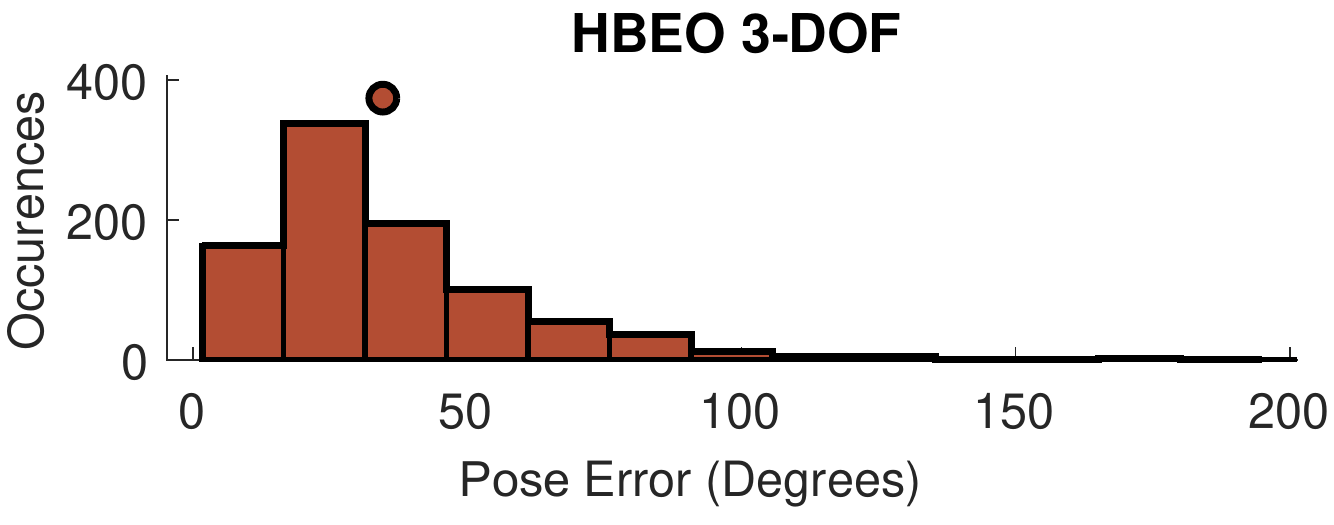}
\end{subfigure}%
\begin{subfigure}[t]{0.32\textwidth}
\includegraphics[width =\textwidth]{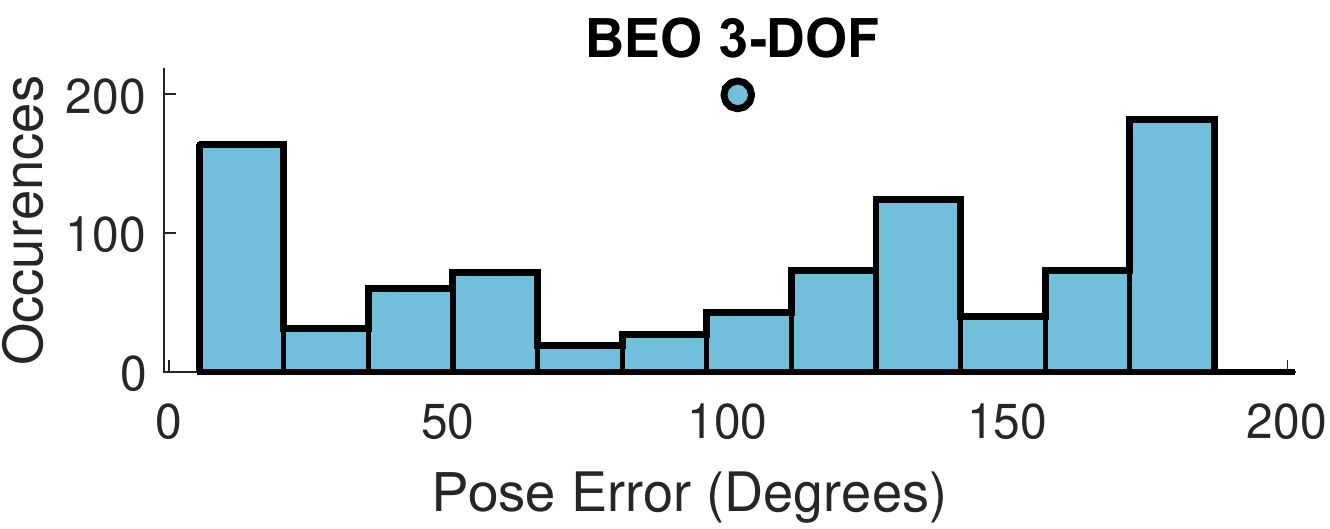}
\end{subfigure}%
\begin{subfigure}[t]{0.32\textwidth}
\includegraphics[width =\textwidth]{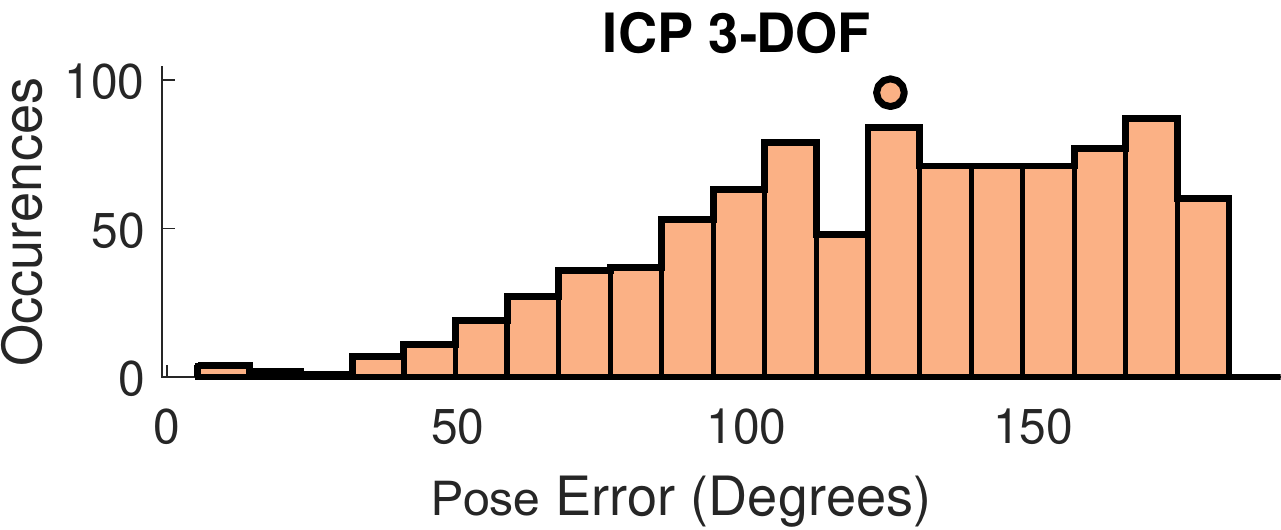}
\end{subfigure}%
\caption{Pose estimation error in 3-DOF for HBEOs, BEOs, and an ICP baseline. Mean error is indicated by circular dots.}
\label{fig:poseEstimationPlots}
\end{figure*}

We then generated roughly 7 million synthetic depth images of size $320$ by $240$ from the objects in our training set by sampling multiple random viewpoints from each of the $3991$ training objects. The ground truth subspace projection for each training object was obtained using equation \eqref{eq:project_complete} and fed to HBEONet during training\footnote{HBEONet required roughly 2 training epochs (16 hours on a single Nvidia GTX1070 GPU) to converge and was implemented using TensorFlow 1.5. The encoded and compressed depth-image dataset required roughly 200GB of storage space.} along with the true pose and class of the object depicted in each depth image.

We compared HBEOs to vanilla BEOs as well as a baseline end-to-end deep method (3DShapeNets). An apples-to-apples comparison here is somewhat difficult; HBEOs, by their very nature, reason over possible poses due to their training regime while 3DShapeNets do not. Furthermore, BEO results in 3-DOF for combined classification and pose estimation proved to be computationally infeasible. As a result, we report 3DShapeNets results with known pose and BEO results with both known pose and 1-DOF unknown pose as comparisons to HBEOs full 3-DOF results.

\subsection{Classification}
Despite HBEOs being required to solve a harder problem than 3DShapeNets and BEOs, classification performance was significantly better than both of them, outperforming (with unknown 3-DOF pose) 3DShapeNets and BEOs with known pose. Table \ref{table:partialClassification} illustrates these results; HBEOs operating on data with unknown pose in 3-DOF has less than half of the misclassification rate of BEOs operating on data with only a single unknown pose DOF. One classification advantage HBEOs posess over BEOs is their ability to perform forward projection jointly with classification. Because BEOs first project into the learned subspace and then classify objects, small details which do not significantly affect shape, but are germane to determining object type, may be missed. This is particularly evident in the case of disambiguating desks, dressers, and nightstands: three classes which have similar overall shape and where small details are important for determining class. Because HBEOs learn to perform classification and subspace projection jointly, they perform much better in this scenario.

\subsection{Pose Estimation}
We evaluate the pose estimation performance of HBEOs by comparing with BEOs and an ICP baseline (see Figure \ref{fig:poseEstimationPlots}). In the HBEO case, class, pose, and 3D geometry are estimated jointly as described in the preceding sections. Due to performance constraints, it was infeasable to directly compare to BEOs. Instead, we provided BEOs with the ground truth class and only required the BEO pipeline to estimate pose and 3D shape. The ICP baseline was also provided with the ground truth class and attempted to estimate pose by aligning the partial input with the mean training object in that class. Despite not having access to the input query's class, HBEOs significantly outperformed both BEOs and the baseline method, achieving a mean pose error less than half that of BEOs. Part of the advantage HBEOs enjoy over BEOs is their direct prediction of pose; because BEOs employ pose estimation by search, they can only sample candidate poses at a relatively course resolution in 3-DOF due to computational constraints, even with known class. HBEOs do not suffer from this drawback as they predict pose parameters directly in a single inference step.

\subsection{3D Completion}
\begin{figure*}[ht!]
\centering
\begin{subfigure}[t]{0.24\textwidth}
\includegraphics[width =\textwidth]{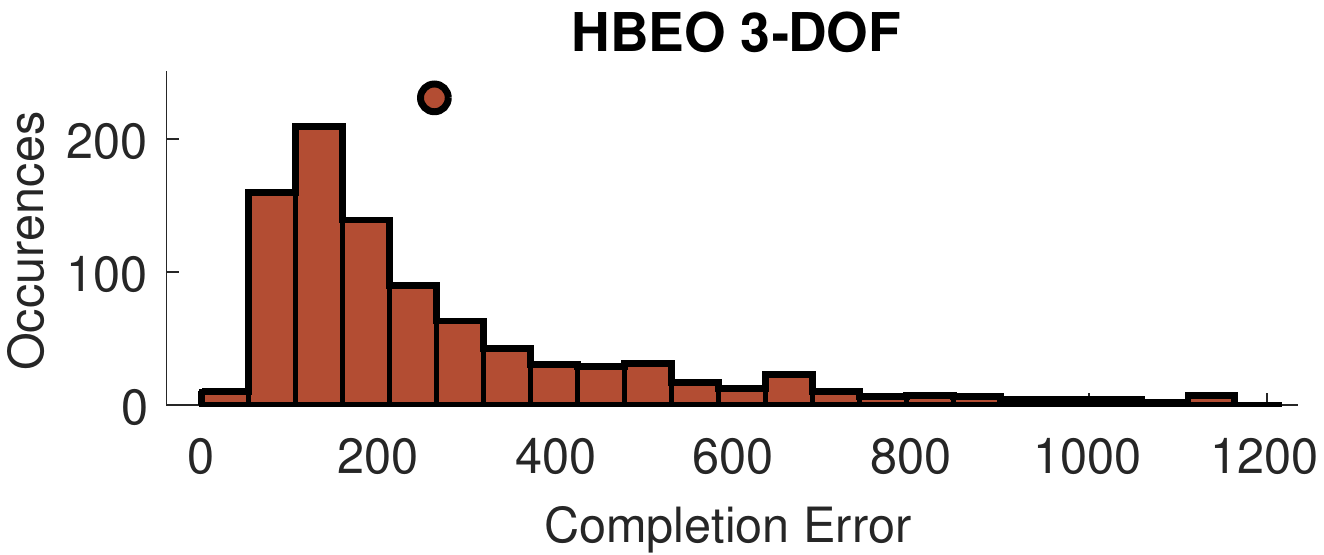}
\end{subfigure}%
\begin{subfigure}[t]{0.24\textwidth}
\includegraphics[width =\textwidth]{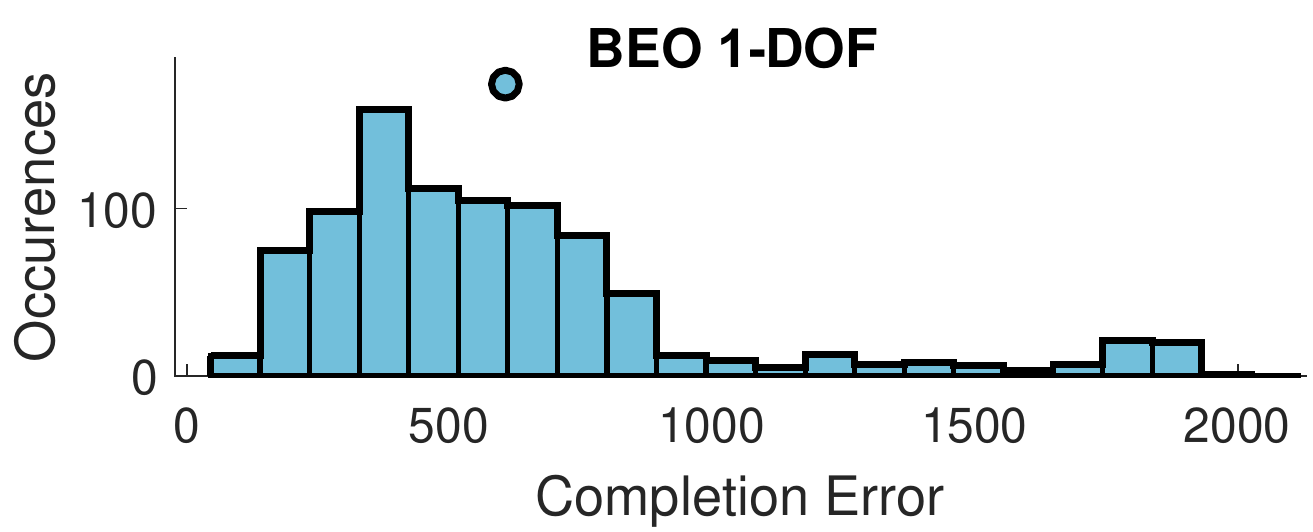}
\end{subfigure}%
\begin{subfigure}[t]{0.24\textwidth}
\includegraphics[width =\textwidth]{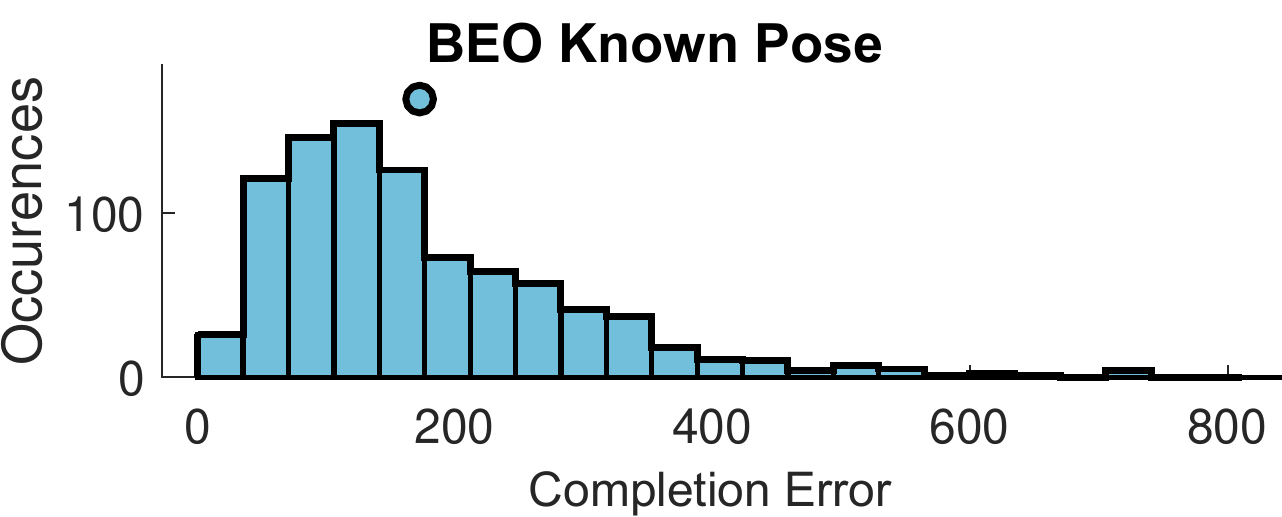}
\end{subfigure}%
\begin{subfigure}[t]{0.24\textwidth}
\includegraphics[width =\textwidth]{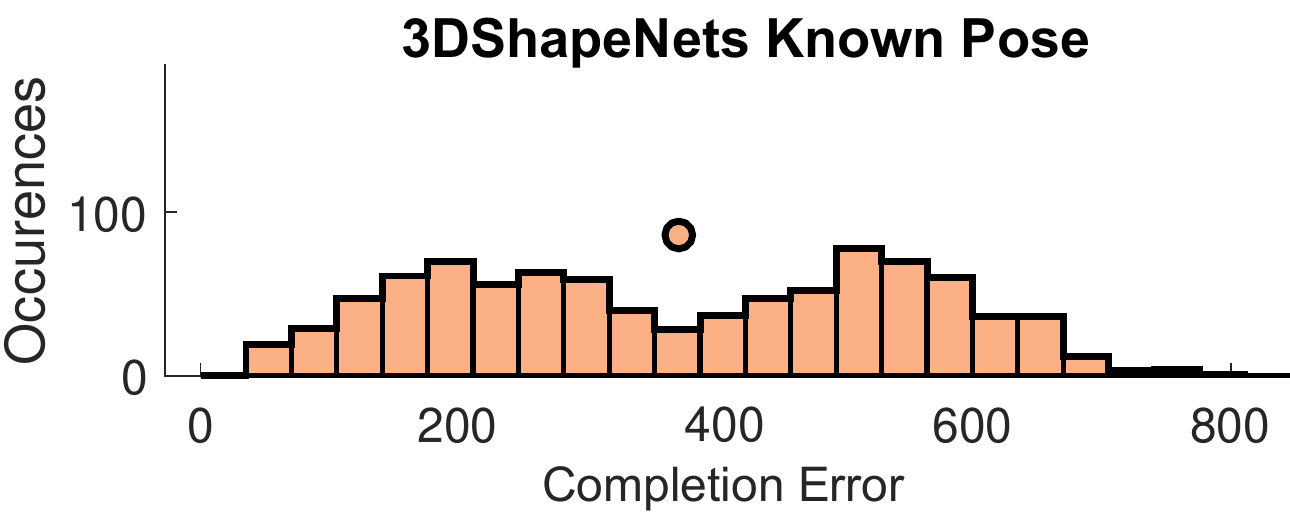}
\end{subfigure}%
\caption{3D completion error using the Euclidean Distance Metric defined in Equation \eqref{eq:reconstScore}.}
\label{fig:completion_performance}
\end{figure*}
\begin{figure*}[ht!]
\centering
\includegraphics[width= \textwidth]{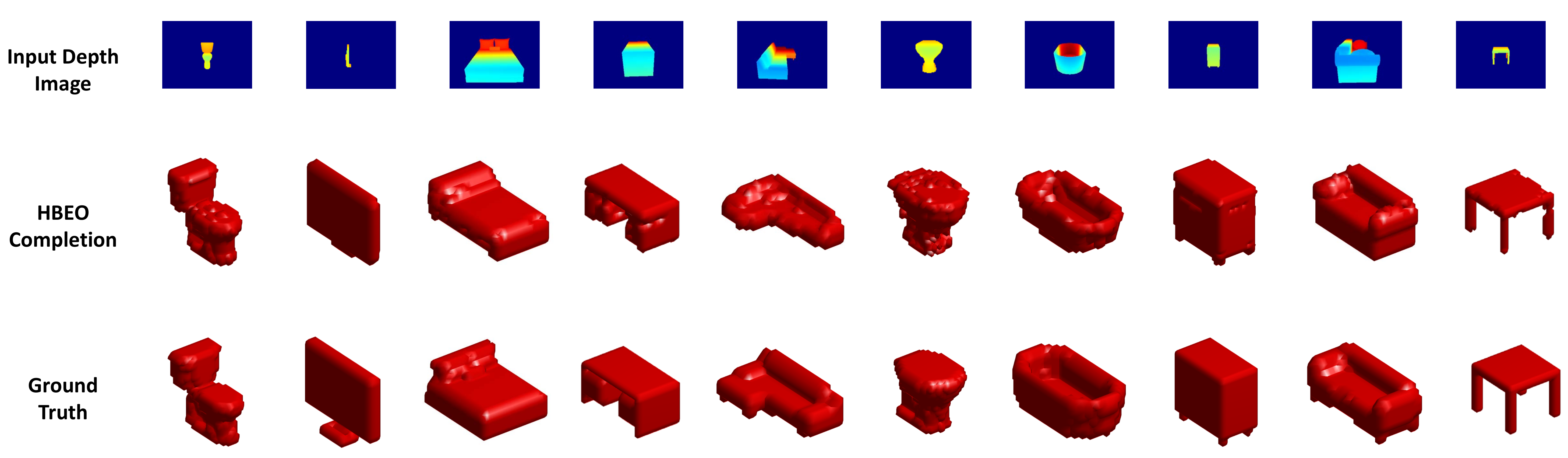}
\caption{Sample completions from the ModelNet10 test set.}
\label{fig:completion_example}
\end{figure*}
While classification accuracy and angular pose error are relatively straightforward to evaluate, 3D completion error is slightly more difficult. The naive approach of measuring completion error is simply \begin{equation}
e(\mathbf{o}, \mathbf{\hat{o}}) = 1-\frac{||\mathbf{o} - \mathbf{\hat{o}}||_2}{|\mathbf{o}|},
\label{eq:reconstScore}
\end{equation}
however this metric is extremely sensitive to misalignment. Because it considers voxel-wise differences between the completed object and ground truth, a completion which is offset slightly but otherwise accurate may still receive a poor score. We thus follow \citet{Burchfiel-RSS-17} in using a more robust metric by extracting the (unsigned) Euclidean Distance Transform (EDT) \cite{1177156} from both the completed objects and the target objects. The EDT creates a distance field where locations in the field that lie on or inside an object have a value of 0 and locations outside of the object receive a value corresponding to the distance of the closest on-object point. Given $\mathbf{D}$ and $\mathbf{\hat{D}}$ denoting the EDT of the target and estimated objects, we can construct a more robust completion score:
\begin{equation}
e'(\mathbf{o}, \mathbf{\hat{o}^r}) = ||\mathbf{D} - \mathbf{\hat{D}}||_2.
\label{eq:dist}
\end{equation}
This EDT-based metric is sensitive to shape change but far less sensitive to small misalignment than equation \eqref{eq:reconstScore}.

Figure \ref{fig:completion_performance} illustrates completion performance of HBEOs relative to BEOs and 3DShapeNets. HBEOs, when performing inference directly from depth images with unknown 3-DOF pose, perform competitively with BEOs operating on perfectly voxelized and aligned input with known pose and significantly outperforms 3DShapeNets (with known pose) and BEOs (with unknown pose in a single DOF). Figure \ref{fig:completion_example} illustrates several example HBEO object completions. In our experience, BEOs become brittle when the query object has unknown pose due to the alignment sensitivity of voxel representations. A small error in estimated pose causes a voxel representation to change dramatically while depth images change in a much smoother fashion. Because of this phenomenon, HBEO completion performance is far more robust to pose estimation errors. 

\subsection{Inference Runtime Performance}
\begin{table}[ht!]
  \begin{tabular}{lr}  
  \toprule
    3DShapeNets \cite{wu20153d} (Known Pose) & $3.57$s\\
    BEO \cite{Burchfiel-RSS-17} (Known Pose) &$1.13$s\\
    BEO \cite{Burchfiel-RSS-17} (1-DOF Pose) &$672.97$s\\
    BEO \cite{Burchfiel-RSS-17} (3-DOF Pose) &$3529.88$s\\
    HBEO (3-DOF Pose) &$0.01$s\\
    \bottomrule
  \end{tabular}
  \caption{Mean runtime performance.}
\label{table:timing}
\end{table}
Comparing timing performance between methods can be difficult; differences in programming language and hardware can affect methods differently. For our testing, HBEOs were implemented in Python (with HBEONet trained using TensorFlow) while BEOs and 3DShapeNets are both implemented in Matlab. As a result, direct comparison of runtimes should be taken with a grain of salt, and small (i.e. 2x or 3x) speed differences between algorithms are not necessarily meaningful in this context. Furthermore, the HBEONet portion of HBEOs is fully GPU accelerated while portions of 3DShapeNets and BEOs are not. Nevertheless, large speed differences (an order of magnitude or more) do highlight gross computational efficiency differences between approaches. Table \ref{table:timing} shows mean run-time performance of 3DShapeNets (for inference on objects of known pose) as well as BEOs and HBEOs. Critically, because HBEOs perform inference in a single shot, they are able to estimate pose in 3-DOF without incurring the large speed penalty that the BEO approach of pose estimation by search produces. While the speed of BEOs can be tweaked to some degree (by adjusting the coarseness of the pose discretization), 
HBEOs are orders of magnitude faster in the full 3-DOF setting. Indeed, HBEOs are fast enough for realtime use while BEOs---in 1-DOF or 3-DOF---are not (without access to a cluster of machines).\footnote{All algorithms were evaluated on a 4-core Intel CPU with 32GB of RAM coupled with an Nvidia GTX1070 GPU.}

\section{Discussion}
HBEOs are significantly faster than preceding methods, offer substantially improved performance, and are much more readily applicable to real problems because they perform inference directly from depth-images. A key property of HBEOs is combining subspace methods with deep learning, learning a low-dimensional subspace into which regression is performed. Because the space of all possible 3D structures is vastly larger than the space of reasonable 3D objects, operating in this subspace dramatically shrinks the size of the inference task. Compared to purely linear methods (such as BEOs) HBEO's use of convolutional nonlinear inference allows it to be far more expressive and less brittle to misalignment or pose-estimation errors.

As robot hardware becomes more adept, perception increasingly bottlenecks the environments our robots can operate in. Particularly in the domain of household robots, where environments exhibit huge variation and are extremely unstructured, we must continue to develop algorithms that allow robots to reason about previously unencountered objects in meaningful ways. To this end, HBEOs constitute a significant step forward for joint 3D object completion, classification, and pose estimation---three critical low-level perceptual tasks in robotics.

\section{Acknowledgments}

This research was supported in part by DARPA under agreement number D15AP00104. The U.S. Government is authorized to reproduce and distribute reprints for Governmental purposes notwithstanding any copyright notation thereon. The content is solely the responsibility of the authors and does not necessarily represent the official views of DARPA.

\bibliographystyle{plainnat}
\bibliography{references}

\begin{thebibliography}{35}
\providecommand{\natexlab}[1]{#1}
\providecommand{\url}[1]{\texttt{#1}}
\expandafter\ifx\csname urlstyle\endcsname\relax
  \providecommand{\doi}[1]{doi: #1}\else
  \providecommand{\doi}{doi: \begingroup \urlstyle{rm}\Url}\fi

\bibitem[Attene(2010)]{attene2010lightweight}
M.~Attene.
\newblock A lightweight approach to repairing digitized polygon meshes.
\newblock \emph{The Visual Computer}, 26:\penalty0 1393--1406, 2010.

\bibitem[Bai et~al.(2016)Bai, Bai, Zhou, Zhang, and Jan~Latecki]{bai2016gift}
S.~Bai, X.~Bai, Z.~Zhou, Z.~Zhang, and L.~Jan~Latecki.
\newblock Gift: A real-time and scalable {3D} shape search engine.
\newblock In \emph{Computer Vision and Pattern Recognition}, 2016.

\bibitem[Bergamo and Torresani(2010)]{NIPS2010_4064}
A.~Bergamo and L.~Torresani.
\newblock Exploiting weakly-labeled web images to improve object
  classification: a domain adaptation approach.
\newblock In \emph{Advances in Neural Information Processing Systems}, pages
  181--189, 2010.

\bibitem[Burchfiel and Konidaris(2017)]{Burchfiel-RSS-17}
B.~Burchfiel and G.~Konidaris.
\newblock Bayesian {E}igenobjects: A unified framework for {3D} robot
  perception.
\newblock In \emph{Robotics: Science and Systems}, 2017.

\bibitem[{C. M. Bishop}(1999{\natexlab{a}})]{bishop1999variational}
{C. M. Bishop}.
\newblock Variational principal components.
\newblock In \emph{International Conference on Artificial Neural Networks},
  pages 509–--514, 1999{\natexlab{a}}.

\bibitem[{C. M. Bishop}(1999{\natexlab{b}})]{bpca1}
{C. M. Bishop}.
\newblock Bayesian {PCA}.
\newblock In \emph{Advances in Neural Information Processing Systems}, pages
  382--388, 1999{\natexlab{b}}.

\bibitem[{C. R. Maurer, Jr., R. Qi, and V. Raghavan}(2003)]{1177156}
{C. R. Maurer, Jr., R. Qi, and V. Raghavan}.
\newblock A linear time algorithm for computing exact {Euclidean} distance
  transforms of binary images in arbitrary dimensions.
\newblock \emph{Pattern Analysis and Machine Intelligence}, 25:\penalty0
  265--270, 2003.

\bibitem[Carbone(2012)]{carbone2012grasping}
G.~Carbone.
\newblock \emph{Grasping in robotics}, volume~10 of \emph{Mechanisms and
  Machine Science}.
\newblock Springer, 2012.

\bibitem[{D. Huber, A. Kapuria, R. Donamukkala, and M. Hebert}(2004)]{3D_Parts}
{D. Huber, A. Kapuria, R. Donamukkala, and M. Hebert}.
\newblock Parts-based {3D} object classification.
\newblock In \emph{Computer Vision and Pattern Recognition}, volume~2, pages
  82--89, 2004.

\bibitem[Dai et~al.(2017)Dai, Qi, and Nie{\ss}ner]{dai2017complete}
A.~Dai, C.~Qi, and M.~Nie{\ss}ner.
\newblock Shape completion using 3d-encoder-predictor {CNNs} and shape
  synthesis.
\newblock In \emph{Computer Vision and Pattern Recognition}, 2017.

\bibitem[Elhoseiny et~al.(2015)Elhoseiny, El-Gaaly, Bakry, and
  Elgammal]{elhoseiny2015convolutional}
M.~Elhoseiny, T.~El-Gaaly, A.~Bakry, and A.~Elgammal.
\newblock Convolutional models for joint object categorization and pose
  estimation.
\newblock \emph{arXiv:1511.05175}, 2015.

\bibitem[Gehler and Nowozin(2009)]{gehler2009feature}
P.~Gehler and S.~Nowozin.
\newblock On feature combination for multiclass object classification.
\newblock In \emph{International Conference on Computer Vision}, pages
  221--228, 2009.

\bibitem[Hegde and Zadeh(2016)]{hegde2016fusionnet}
V.~Hegde and R.~Zadeh.
\newblock Fusionnet: {3D} object classification using multiple data
  representations.
\newblock \emph{rXiv:1607.05695}, 2016.

\bibitem[{J. Glover, R. Rusu, and G. Bradski}(2011)]{Glover-RSS-11}
{J. Glover, R. Rusu, and G. Bradski}.
\newblock {Monte Carlo} pose estimation with quaternion kernels and the
  {B}ingham distribution.
\newblock In \emph{Robotics: Science and Systems}, 2011.

\bibitem[Krizhevsky et~al.(2012)Krizhevsky, Sutskever, and
  Hinton]{krizhevsky2012imagenet}
A.~Krizhevsky, I.~Sutskever, and G.~Hinton.
\newblock Imagenet classification with deep convolutional neural networks.
\newblock In \emph{Advances in neural information processing systems}, pages
  1097--1105, 2012.

\bibitem[LeCun and Bengio(1995)]{lecun1995convolutional}
Y.~LeCun and Y.~Bengio.
\newblock Convolutional networks for images, speech, and time series.
\newblock \emph{The handbook of brain theory and neural networks},
  3361\penalty0 (10):\penalty0 1995, 1995.

\bibitem[Li et~al.(2015)Li, Dai, Guibas, and
  Nie{\ss}ner]{guibas_database_objects2}
Y.~Li, A.~Dai, L.~Guibas, and M.~Nie{\ss}ner.
\newblock Database-assisted object retrieval for real-time {3D} reconstruction.
\newblock In \emph{Computer Graphics Forum}, volume~34, pages 435--446, 2015.

\bibitem[{M. E. Tipping and C. M. Bishop}(1999)]{PPCA}
{M. E. Tipping and C. M. Bishop}.
\newblock Probabilistic principal component analysis.
\newblock \emph{Journal of the Royal Statistical Society. Series B (Statistical
  Methodology)}, 61:\penalty0 611--622, 1999.

\bibitem[Marini et~al.(2006)Marini, Biasotti, and
  Falcidieno]{marini2006partial}
S.~Marini, S.~Biasotti, and B.~Falcidieno.
\newblock Partial matching by structural descriptors.
\newblock In \emph{Content-Based Retrieval}, 2006.

\bibitem[Maturana and Scherer(2015)]{maturana2015voxnet}
D.~Maturana and S.~Scherer.
\newblock Voxnet: A {3D} convolutional neural network for real-time object
  recognition.
\newblock In \emph{Intelligent Robots and Systems}, pages 922--928, 2015.

\bibitem[{P. J. Besl and N. D. McKay}(1992)]{ICP}
{P. J. Besl and N. D. McKay}.
\newblock Method for registration of {3-D} shapes.
\newblock \emph{Pattern Analysis and Machine Intelligence}, 14:\penalty0
  239--256, 1992.

\bibitem[Qi et~al.(2016)Qi, Su, Niessner, Dai, Yan, and Guibas]{Qi_2016_CVPR}
C.~Qi, H.~Su, M.~Niessner, A.~Dai, M.~Yan, and L.~Guibas.
\newblock Volumetric and multi-view {CNNs} for object classification on {3D}
  data.
\newblock In \emph{Computer Vision and Pattern Recognition}, 2016.

\bibitem[Rusu and Cousins(2011)]{Rusu20113DIH}
R.~Rusu and S.~Cousins.
\newblock {3D} is here: Point cloud library (pcl).
\newblock \emph{International Conference on Robotics and Automation}, pages
  1--4, 2011.

\bibitem[Schiebener et~al.(2016)Schiebener, Schmidt, Vahrenkamp, and
  Asfour]{schiebener2016heuristic}
D.~Schiebener, A.~Schmidt, N.~Vahrenkamp, and T.~Asfour.
\newblock Heuristic {3D} object shape completion based on symmetry and scene
  context.
\newblock In \emph{Intelligent Robots and Systems}, pages 74--81, 2016.

\bibitem[Shi et~al.(2015)Shi, Bai, Zhou, and Bai]{7273863}
B.~Shi, S.~Bai, Z.~Zhou, and X.~Bai.
\newblock Deeppano: Deep panoramic representation for {3-D} shape recognition.
\newblock \emph{Signal Processing Letters}, 22\penalty0 (12):\penalty0
  2339--2343, 2015.

\bibitem[Soltani et~al.(2017)Soltani, Huang, Wu, Kulkarni, and
  Tenenbaum]{Soltani2017Synthesizing3S}
A.~Soltani, H.~Huang, J.~Wu, T.~Kulkarni, and J.~Tenenbaum.
\newblock Synthesizing {3D} shapes via modeling multi-view depth maps and
  silhouettes with deep generative networks.
\newblock \emph{Computer Vision and Pattern Recognition}, pages 2511--2519,
  2017.

\bibitem[Song and Xiao(2016)]{Song_2016_CVPR}
S.~Song and J.~Xiao.
\newblock Deep sliding shapes for amodal {3D} object detection in {RGB-D}
  images.
\newblock In \emph{Computer Vision and Pattern Recognition}, 2016.

\bibitem[Su et~al.(2015)Su, Maji, Kalogerakis, and Learned-Miller]{su2015multi}
H.~Su, S.~Maji, E.~Kalogerakis, and E.~Learned-Miller.
\newblock Multi-view convolutional neural networks for {3D} shape recognition.
\newblock In \emph{International Conference on Computer Vision}, pages
  945--953, 2015.

\bibitem[Tulsiani and Malik(2015)]{tulsiani2015viewpoints}
S.~Tulsiani and J.~Malik.
\newblock Viewpoints and keypoints.
\newblock In \emph{Computer Vision and Pattern Recognition}, pages 1510--1519,
  2015.

\bibitem[Tulsiani et~al.(2015)Tulsiani, Kar, Huang, Carreira, and
  Malik]{DBLP:journals/corr/TulsianiKHCM15}
S.~Tulsiani, A.~Kar, Q.~Huang, J.~Carreira, and J.~Malik.
\newblock Shape and symmetry induction for {3D} objects.
\newblock \emph{CoRR}, abs/1511.07845, 2015.

\bibitem[Varley et~al.(2017)Varley, DeChant, Richardson, Ruales, and
  Allen]{varlay2017}
J.~Varley, C.~DeChant, A.~Richardson, J.~Ruales, and P.~Allen.
\newblock Shape completion enabled robotic grasping.
\newblock In \emph{Intelligent Robots and Systems}, pages 2442--2447, 2017.

\bibitem[Wu et~al.(2016)Wu, Zhang, Xue, Freeman, and Tenenbaum]{wu2016learning}
J.~Wu, C.~Zhang, T.~Xue, B.~Freeman, and J.~Tenenbaum.
\newblock Learning a probabilistic latent space of object shapes via {3D}
  generative-adversarial modeling.
\newblock In \emph{Advances in Neural Information Processing Systems}, pages
  82--90, 2016.

\bibitem[Wu et~al.(2015)Wu, Song, Khosla, Yu, Zhang, Tang, and Xiao]{wu20153d}
Z.~Wu, S.~Song, A.~Khosla, F.~Yu, L.~Zhang, X.~Tang, and J.~Xiao.
\newblock {3D} shapenets: A deep representation for volumetric shapes.
\newblock In \emph{Computer Vision and Pattern Recognition}, pages 1912--1920,
  2015.

\bibitem[{Y. Kim, N. J. Mitra, D. M. Yan, and L.
  Guibas}(2012)]{guibas_database_objects}
{Y. Kim, N. J. Mitra, D. M. Yan, and L. Guibas}.
\newblock Acquiring {3D} indoor environments with variability and repetition.
\newblock \emph{ACM Transactions on Graphics}, 31:\penalty0 138:1--138:11,
  2012.

\bibitem[{Y. Kim, N. J. Mitra, Q. Huang, and L.
  Guibas}(2013)]{guided3DScanning}
{Y. Kim, N. J. Mitra, Q. Huang, and L. Guibas}.
\newblock Guided real-time scanning of indoor objects.
\newblock In \emph{Computer Graphics Forum}, volume~32, pages 177--186, 2013.

\end{thebibliography}

\end{document}